\documentclass[conference]{IEEEtran}
\IEEEoverridecommandlockouts
\usepackage{cite}
\usepackage{amsmath,amssymb,amsfonts}
\usepackage{algorithmic}
\usepackage{graphicx}
\usepackage{textcomp}
\usepackage{xcolor}
\def\BibTeX{{\rm B\kern-.05em{\sc i\kern-.025em b}\kern-.08em
    T\kern-.1667em\lower.7ex\hbox{E}\kern-.125emX}}

\usepackage{tabularx}
\usepackage{placeins}
\usepackage{float}
\usepackage{booktabs} 
\usepackage{array} 
\usepackage{multirow} 
\newcolumntype{V}{!{\vrule width 1.5pt}}
\usepackage{amsmath}
\usepackage{amssymb}
\usepackage{pifont}
\usepackage[table]{xcolor}
\usepackage{booktabs} 
\usepackage{multirow} 
\usepackage{array}    
\usepackage{xcolor}   
\usepackage{pifont}
\usepackage{bbding}  
\usepackage{enumitem}
\usepackage{amsmath, amssymb, mathrsfs, bm}
\usepackage{graphicx, subcaption}
\usepackage{tabularx}
\usepackage{stfloats}
\newcolumntype{Y}{>{\raggedright\arraybackslash}X}

\newcommand{\chance}{\mathfrak{C}}
\newcommand{\accuracy}{\mathsf{A}}
\newcommand{\score}{\mathscr{S}}

\definecolor{bluecheck}{RGB}{0, 0, 255} 
\definecolor{redcross}{RGB}{255, 0, 0}   

\begin{document}

\title{MVPBench: A Multi-Video Perception Evaluation Benchmark for Multi-Modal Video Understanding}

\author{\IEEEauthorblockN{1\textsuperscript{st} Purui Bai}
\IEEEauthorblockA{\textit{MAIS \& NLPR, Institute of Automation} \\
\textit{Chinese Academy of Sciences}\\
Beijing, China \\
purui.bai@cripac.ia.ac.cn}
\and
\IEEEauthorblockN{2\textsuperscript{nd} Tao Wu}
\IEEEauthorblockA{\textit{SIST} \\
\textit{ShanghaiTech University}\\
Shanghai, China \\
wutao2022@shanghaitech.edu.cn}
\and
\IEEEauthorblockN{3\textsuperscript{rd} Jiayang Sun}
\IEEEauthorblockA{\textit{MAIS \& NLPR, Institute of Automation} \\
\textit{Chinese Academy of Sciences}\\
Beijing, China \\
sunjiayang2025@ia.ac.cn}
\and
\IEEEauthorblockN{4\textsuperscript{th} Xinyue Liu}
\IEEEauthorblockA{\textit{MAIS \& NLPR, Institute of Automation} \\
\textit{Chinese Academy of Sciences}\\
Beijing, China \\
liuxinyue2025@ia.ac.cn}
\and
\IEEEauthorblockN{5\textsuperscript{th} Huaibo Huang}
\IEEEauthorblockA{\textit{MAIS \& NLPR, Institute of Automation} \\
\textit{Chinese Academy of Sciences}\\
Beijing, China \\
huaibo.huang@cripac.ia.ac.cn}
\and
\IEEEauthorblockN{6\textsuperscript{th} Ran He}
\IEEEauthorblockA{\textit{MAIS \& NLPR, Institute of Automation} \\
\textit{Chinese Academy of Sciences}\\
Beijing, China \\
ran.he@cripac.ia.ac.cn}
}

\maketitle

\begin{abstract}
The rapid progress of Large Language Models (LLMs) has spurred growing interest in Multi-modal LLMs (MLLMs) and motivated the development of benchmarks to evaluate their perceptual and comprehension abilities. Existing benchmarks, however, are limited to static images or single videos, overlooking the complex interactions across multiple videos. To address this gap, we introduce the \textbf{M}ulti-\textbf{V}ideo \textbf{P}erception Evaluation \textbf{Bench}mark (\textbf{MVPBench}), a new benchmark featuring \textbf{14} subtasks across diverse visual domains designed to evaluate models on extracting relevant information from video sequences to make informed decisions. MVPBench includes \textbf{5K} question-answering tests involving \textbf{2.7K} video clips sourced from existing datasets and manually annotated clips. Extensive evaluations reveal that current models struggle to process multi-video inputs effectively, underscoring substantial limitations in their multi-video comprehension. We anticipate MVPBench will drive advancements in multi-video perception.
\end{abstract}


\section{Introduction}

\begin{figure*}[t]
    \centering
    \includegraphics[width=\textwidth]{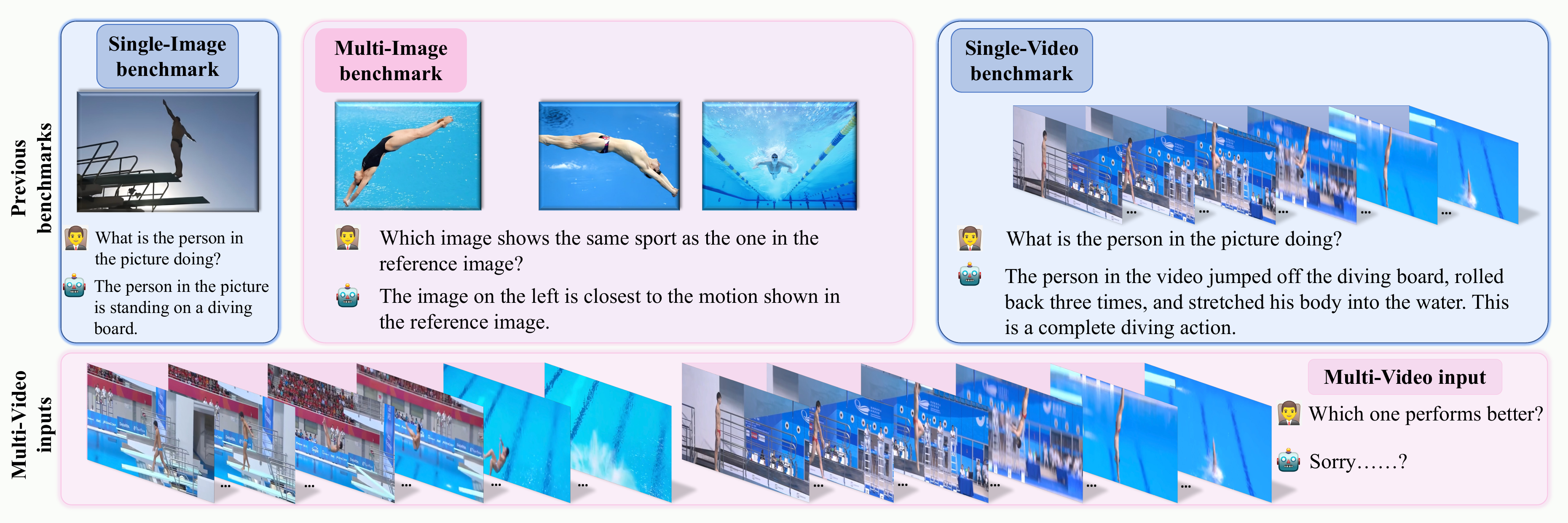}
    \caption{\textbf{Challenges in managing Multi-Video inputs.} Previous benchmarks are primarily designed to address straightforward visual text problems, such as describing the content and associations of static images. While video description provides additional information to enhance the detail of content descriptions, it remains challenging to effectively manage comparisons between multiple related videos.}
    \label{fig:teaser}
\end{figure*}

With the advancement of Large Language Models (LLMs) \cite{openai2024gpt, touvron2023llama2}, there is a growing interest in the application within the domain of visual modalities, further leading to the development of Multi-modal Large Language Models (MLLMs) \cite{liu2024visual, bai2023qwen}. As multi-modal application technologies have progressed, corresponding evaluation benchmarks have also been developed, which evolved from single images \cite{xu2023lvlm, liu2025mmbench}, to multiple images \cite{li2024seed, fu2024blink}, and eventually to video modalities \cite{li2024seed, li2024mvbench, fu2024video}.
However, a significant gap exists in current MLLMs’ evaluation benchmarks, in that they do not assess the models' capabilities in handling multiple dynamic video inputs. As illustrated in Fig. \ref{fig:teaser}, these evaluation frameworks fall short in measuring the processing abilities of models with real-world scenarios input.

To address the aforementioned challenges, we propose the Multi-Video Perception Evaluation Benchmark (MVPBench) for Multi-modal Video Understanding. This benchmark provides a comprehensive assessment of MLLMs’ capabilities in processing and understanding multi-video inputs. To the best of our knowledge, MVPBench represents the first benchmark in the MLLM research domain explicitly designed to evaluate multi-video comprehension.

\begin{figure}[t]
    \centering
    \includegraphics[width=\columnwidth]{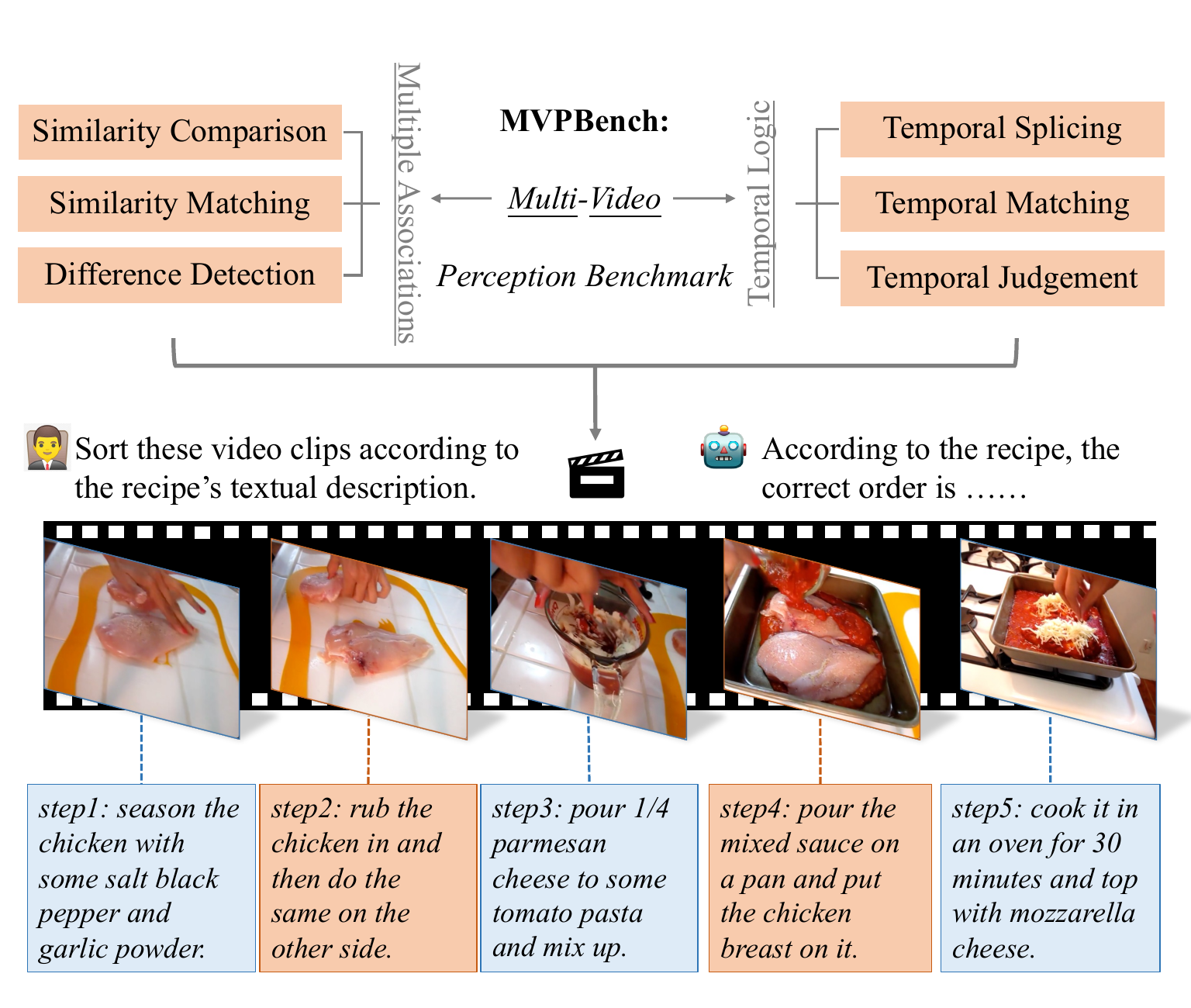}
    \caption{\textbf{Design principles of MVPBench.} A key feature of MVPBench is its ability to consider both the multiplicity of evaluation inputs and the temporal characteristics of evaluation videos.}
    \label{fig:2}
\end{figure}

MVPBench includes a diverse set of \textbf{14} subtasks that evaluate the model's capabilities across various dimensions. These tasks range from basic to advanced levels, covering a variety of question-answering formats from low-level pattern recognition to high-level semantic interpretation, thereby imposing rigorous demands on model performance across perceptual and cognitive dimensions. During the design process, these tasks adhere to the design principles outlined in Fig. \ref{fig:2}, which emphasize the consideration of both the multiplicity of evaluation inputs and the temporal aspects of evaluation videos. As benchmark designers, our core principle is to enable a comprehensive evaluation of MLLMs. Accordingly, our framework is designed to encompass both multi-video processing capabilities and a set of complementary tasks that do not strictly require multiple video inputs, the task effectiveness of which has been demonstrated through repeated validation by prior works. 

\newcolumntype{G}{>{\columncolor{gray!20}}l}
\renewcommand{\arraystretch}{1.2}

{\Large 
\begin{table*}[ht]
\centering
\caption{\textbf{Prompt examples of MVPBench.}}
\resizebox{\textwidth}{!}{%
\begin{tabular}{@{\extracolsep{\fill}}c|c|c|G}
\hline\hline
\multicolumn{1}{c|}{\textbf{Subdomain}} & \multicolumn{1}{c|}{\textbf{Subtask}} & \multicolumn{1}{c|}{\textbf{Source}} & \multicolumn{1}{G}{\textbf{Prompt}} \\ 
\hline\hline
\multirow{2}{*}{\textbf{Temporal Segment}} & Tem w/ & \multirow{2}{*}{YouCook2 \cite{zhou2018towards}} & \textit{Sort these video clips according
to the recipe's textual description. Recipe Description:} \\
& Cap & & \textit{step0: add water to a pan. step1: stir in the pack of sauce. step2: stir and cook until the soup boils} \\
\cline{2-4}
\multirow{2}{*}{\textbf{Splicing}} & Tem w/o & Manual & \textit{Sort the given video clips.} \\
& Cap & Collection & \textit{According to their content or logical sequence.} \\
\hline
\multirow{6}{*}{\textbf{Content Assessment}} & Daily life & \multirow{2}{*}{EPIC-Skills \cite{doughty2018s}} & \textit{Your task is to compare these two clips.} \\
& action evaluation & & \textit{Determine which one has better task and action completion.} \\
\cline{2-4}
& Professional action & \multirow{2}{*}{AQA-7 \cite{parmar2019action}} & \textit{Your task is to compare these two clips.} \\
& evaluation & & \textit{Determine which one has better task and action completion.} \\
\cline{2-4}
& Weather condition & Manual & \textit{Your task is to examine the weather conditions in each video.} \\
& evaluation &  Collection & \textit{Determine which video shows the most severe weather.} \\
\hline
\multirow{2}{*}{\textbf{Video}} & Video clarity & \multirow{2}{*}{CVQAD \cite{antsiferova2022video}} & \textit{Compare these videos based on their clarity and smoothness (video quality).} \\
& evaluation & & \textit{Rank them from highest to lowest quality.} \\
\cline{2-4}
\multirow{2}{*}{\textbf{Quality}} & Video brightness & \multirow{2}{*}{SDSD \cite{wang2021seeing}} & \textit{Identify the video clip among the three given clips.} \\
& evaluation & & \textit{With the lowest brightness and the darkest shooting conditions.} \\
\cline{2-4}
\multirow{2}{*}{\textbf{Assessment}} & Video forensic & Manual & \textit{Compare the three video clips.} \\
& detection &  Collection & \textit{Determine which one is the real video.} \\
\hline
\multirow{2}{*}{\textbf{Video Logic}} & Multiview visual & \multirow{2}{*}{PEV \cite{yonetani2016recognizing}} & \textit{The reference video and the correct candidate video are the corresponding perspectives.} \\
& perception pairing & & \textit{Identify which of the three candidate videos corresponds to the reference video.} \\
\cline{2-4}
\multirow{2}{*}{\textbf{Inference}} & Common sense judgment & Manual & \textit{Please compare the following three video clips.} \\
& of physical laws & Collection & \textit{Identify which one adheres to normal physical laws.} \\
\hline
& Gait recognition & \multirow{2}{*}{CASIA Gait Database \cite{yu2006framework}} & \textit{The reference video and the correct candidate video depict the same person walking.} \\
& matching & & \textit{Determine which of the two candidate videos matches the reference video.} \\
\cline{2-4}
\multirow{2}{*}{\textbf{Similar Video}} & Cinematographic style & Manual & \textit{The reference video and the correct candidate video share the same filming style.} \\
& matching & Collection & \textit{Choose the video that matches the reference video.} \\
\cline{2-4}
\multirow{2}{*}{\textbf{Pairing}} & Olympic sports & Manual & \textit{The reference video and the correct candidate video depict the same sports activity.} \\
& matching & Collection & \textit{Compare the two candidate videos and select the one that matches the reference video.} \\
\cline{2-4}
& Dance style & Manual & \textit{The reference video and the correct candidate video depict the same dance type.} \\
& matching & Collection & \textit{Compare the two candidate videos and select the one that matches the reference video.} \\
\hline\hline
\end{tabular}%
}
\label{tab:1}
\end{table*}
}

To validate the rationality of MVPBench, we conducted evaluations on \textbf{15} state-of-the-art MLLMs of different scales. Results indicate that even the most advanced models achieve an average accuracy of only \textbf{31.10\%}, significantly below human performance at \textbf{88.89\%}. The variation in model performance across different task types highlights MVPBench's capability in pinpointing the limitations of current models in perception tasks. We expect MVPBench to drive the development of MLLMs with enhanced multi-video comprehension capabilities. The complete evaluation code and dataset will be available upon acceptance.

\section{Related Work}

\noindent{\bf Video MLLM.}
In recent years, researchers have expanded visual modalities from static images to dynamic videos, employing innovative technologies such as VideoChat \cite{li2023videochat}, and Valley \cite{luo2023valley} to exploit the substantial potential of LLMs in video understanding tasks. Recent studies, including Qwen2.5-VL \cite{bai2025qwen2} and LLaVA-Video \cite{zhang2024video}, have elevated MLLMs’ capabilities in the video modality through extensive video-text training data pairs.

However, the majority of training data for the video modality remain limited to single video-text pairs, restricting their ability to process multiple videos simultaneously. Currently, there are no established evaluation standards for MLLM’s multi-video understanding capabilities in the research community.

\noindent{\bf Video understanding benchmark.}
Recent advancements in video understanding benchmarks have expanded the evaluation of Video-Language Models (VLMs) across diverse domains and temporal scales. Early benchmarks \cite{jang2017tgif,xu2017video} focused on general video-language understanding, while more recent benchmarks address specialized capabilities, such as MVBench \cite{li2024mvbench} and Video-MME \cite{fu2024video}, featuring videos with multimodal elements (e.g., \textit{subtitles, audio}), EgoSchema \cite{mangalam2023egoschema} and LongVideoBench \cite{wu2024longvideobench} evaluate event- or story-level understanding across extended temporal horizons.

Notably, VideoVista \cite{li2024videovista} incorporates a multi-video comprehension category, namely the Video-Video Relation Reasoning mentioned in their work. However, VideoVista is fundamentally still a conventional benchmark primarily focused on single-video understanding and reasoning. In contrast, our proposed MVPBench is a specialized, holistic, and rigorously designed benchmark which aims to evaluate the diverse perception and reasoning capabilities of MLLMs across multiple video inputs by encompassing various task areas that require joint comprehension and extended reasoning across multiple videos.

\section{MVPBench}
\begin{figure}[h]
    \centering
    \includegraphics[width=0.9\columnwidth]{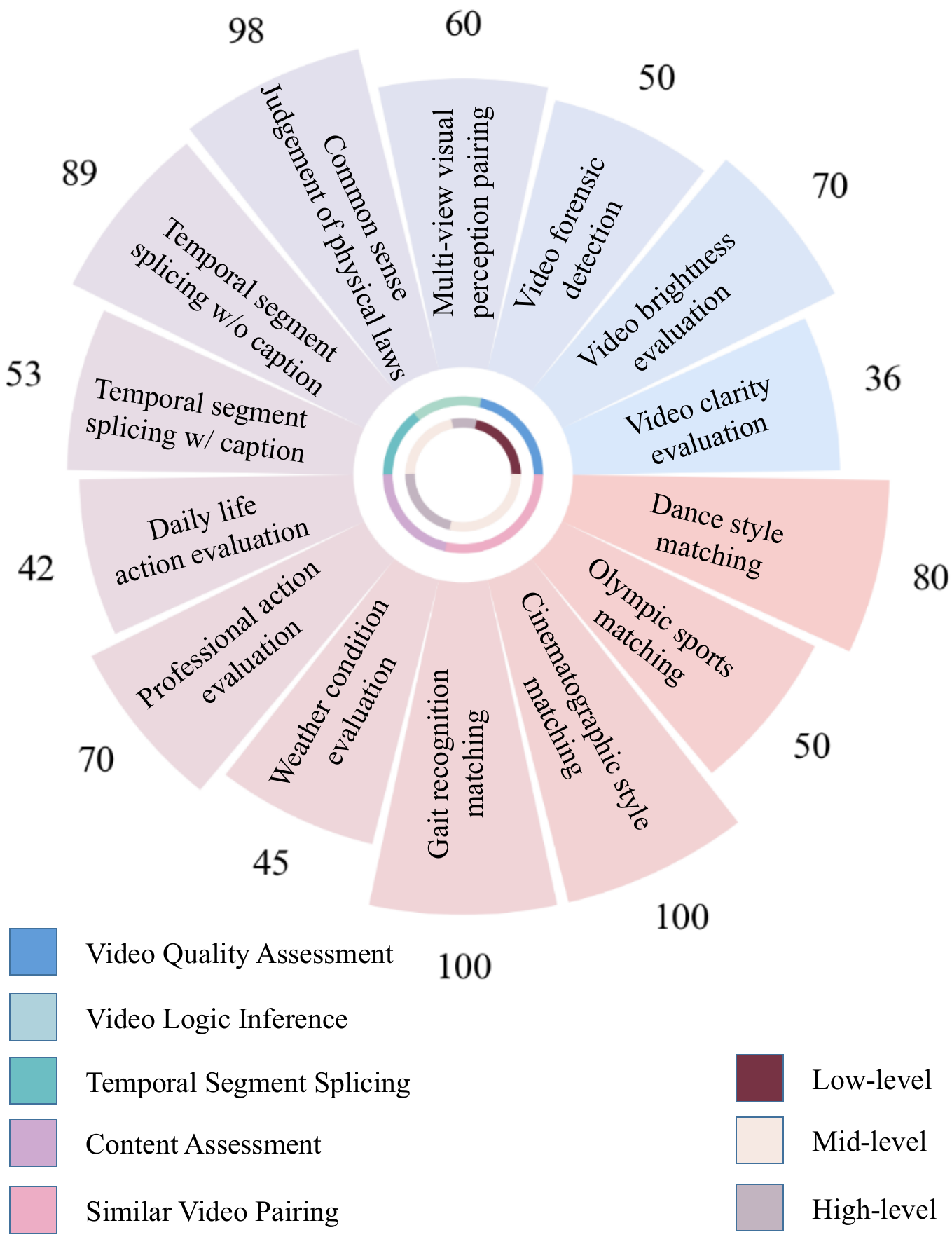}
    \caption{\textbf{Statistics of MVPBench.} The benchmark includes 14 tasks in 5 domains, ranging from low-level pattern comparison to mid-level temporal logic reasoning, and extending to high-level visual content understanding.}
    \label{fig:3}
\end{figure}

MVPBench encompasses a range of visual tasks across different dimensions and levels, with dataset partly derived from existing datasets related to the task, supplemented by a manually curated substantial collection of high-quality video materials. As shown in Tab. \ref{tab:1}, we standardized the prompt design and processing workflow to ensure clear task definitions and to mitigate the influence of extraneous factors such as ambiguous textual information. In total, we contributed 5K visual question answering questions and 2.7K videos. Fig. \ref{fig:3} illustrates the distribution of each task.

\newcommand{\smallcheck}{\scalebox{0.8}{\Checkmark}}
\newcommand{\smallcross}{\scalebox{0.8}{\XSolidBrush}}

\begin{table*}[h]
\centering
\caption{\textbf{The comparison of various benchmarks:} whether the benchmark incorporates videos sourced from diverse open domains (Open.), whether the benchmark incorporates egocentric videos(Ego.), whether the benchmark incorporates complex tasks spanning multiple hierarchical levels of perception(Com.), whether the benchmark incorporates multi-video comprehension category (Multi.) and whether the benchmark incorporates videos synthesized by the generative model(Gen.)}
\renewcommand{\arraystretch}{0.01}
\setlength{\tabcolsep}{1pt} 
\resizebox{\textwidth}{!}{
\begin{tabular}{l|cc|ccccc||c}
\toprule
\textbf{Benchmarks} & \textbf{\#Videos} & \textbf{\#QA Pairs} & \textbf{Open.} & \textbf{Ego.} & \textbf{Com.} & \textbf{Multi.} & \textbf{Gen.} & \textbf{Data source}\\
\midrule
MVBench \cite{li2024mvbench} & 3,641 & 4,000 & \textcolor{bluecheck}{\smallcheck} & \textcolor{bluecheck}{\smallcheck} & \textcolor{bluecheck}{\smallcheck} & \textcolor{redcross}{\smallcross} & \textcolor{redcross}{\smallcross} & existing datasets\\
Video-Bench \cite{ning2023video} & 5,917 & 17,036 & \textcolor{bluecheck}{\smallcheck} & \textcolor{redcross}{\smallcross} & \textcolor{bluecheck}{\smallcheck} & \textcolor{redcross}{\smallcross} & \textcolor{redcross}{\smallcross} & existing datasets\\
EgoSchema \cite{mangalam2023egoschema} & 5,063 & 5,063 & \textcolor{redcross}{\smallcross} & \textcolor{bluecheck}{\smallcheck} & \textcolor{redcross}{\smallcross} & \textcolor{redcross}{\smallcross} & \textcolor{redcross}{\smallcross} & ego-centric video\\
AutoEval-Video \cite{chen2024autoeval} & 327 & 327 & \textcolor{bluecheck}{\smallcheck} & \textcolor{redcross}{\smallcross} & \textcolor{redcross}{\smallcross} & \textcolor{redcross}{\smallcross} & \textcolor{redcross}{\smallcross} & Youtube\\
TempCompass \cite{liu2024tempcompass} & 410 & 7,540 & \textcolor{bluecheck}{\smallcheck} & \textcolor{redcross}{\smallcross} & \textcolor{bluecheck}{\smallcheck} & \textcolor{redcross}{\smallcross} & \textcolor{redcross}{\smallcross} & ShutterStock\\
Video-MME \cite{fu2024video} & 900 & 2,700 & \textcolor{bluecheck}{\smallcheck} & \textcolor{redcross}{\smallcross} & \textcolor{bluecheck}{\smallcheck} & \textcolor{redcross}{\smallcross} & \textcolor{redcross}{\smallcross} & Youtube\\
LongVideoBench \cite{wu2024longvideobench} & 3,763 & 6,678 & \textcolor{bluecheck}{\smallcheck} & \textcolor{redcross}{\smallcross} & \textcolor{bluecheck}{\smallcheck} & \textcolor{redcross}{\smallcross} & \textcolor{redcross}{\smallcross} & web channels\\
VideoVista \cite{li2024videovista} & 894 & 24,906 & \textcolor{bluecheck}{\smallcheck} & \textcolor{redcross}{\smallcross} & \textcolor{bluecheck}{\smallcheck} & \textcolor{bluecheck}{\smallcheck} & \textcolor{redcross}{\smallcross} & YouTube\\
\midrule
MVPBench & 2,774 & 5,050 & \textcolor{bluecheck}{\smallcheck} & \textcolor{bluecheck}{\smallcheck} & \textcolor{bluecheck}{\smallcheck} & \textcolor{bluecheck}{\smallcheck} & \textcolor{bluecheck}{\smallcheck} & web videos, synthetic videos, datasets\\
\bottomrule
\end{tabular}
}
\label{tab:2}
\end{table*}

\subsection{Dataset Collection Process}
\label{sec:3.2}

\begin{figure*}[h!]
    \centering
    \includegraphics[width=\textwidth]{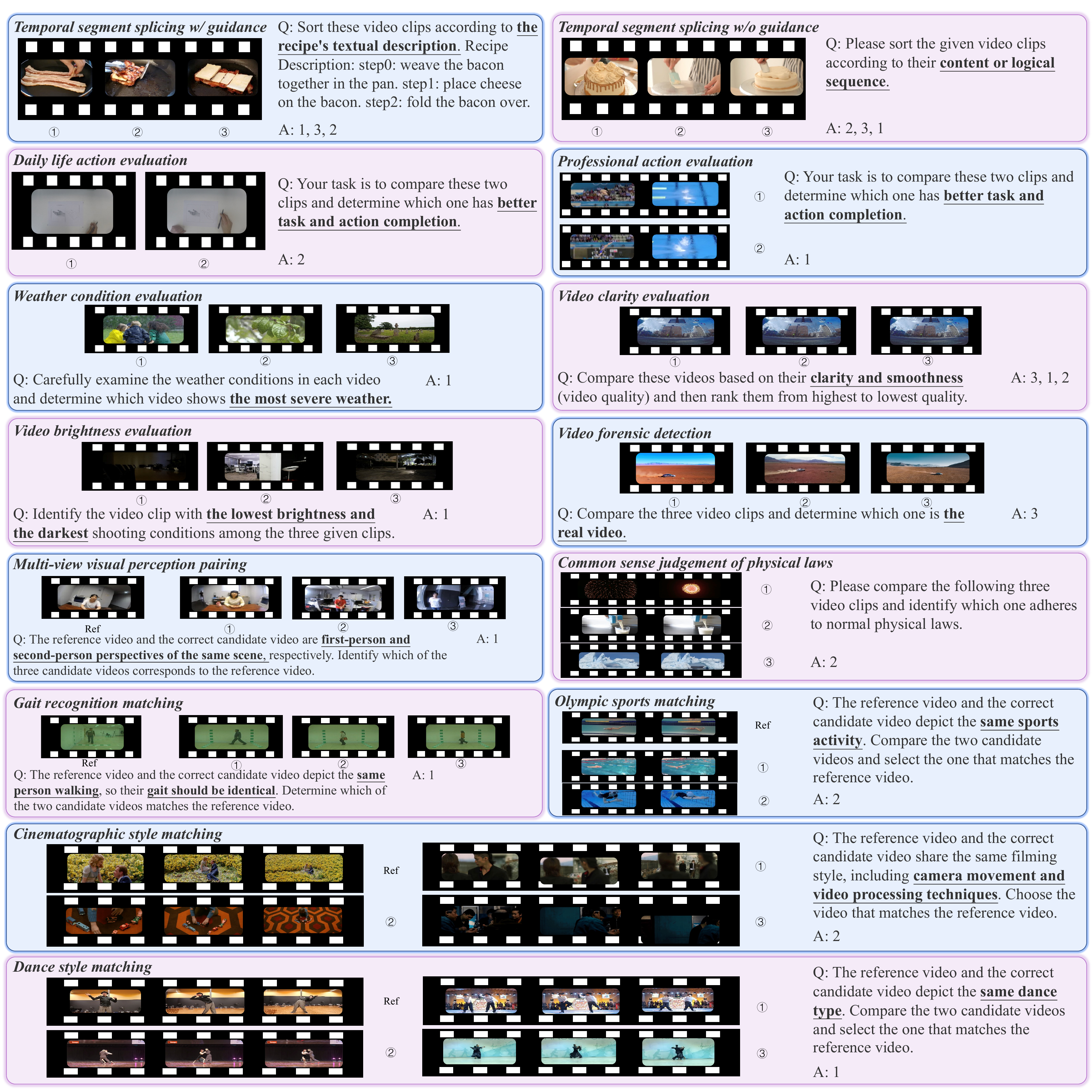}
    \captionsetup{justification=justified} 
    \caption{\textbf{Task introduction of MVPBench}. Zoom in for better view.}
    \label{fig:4}
\end{figure*}

\noindent{\bf Caption-guided Temporal Segment Splicing.} Assesses MLLMs' ability to reconstruct shuffled video sequences leveraging text-visual alignment. Using YouCook2 \cite{zhou2018towards}, we segment videos into recipe-step clips. Given full textual step sequences, models reorder clips based on temporal ordering prompts.

\noindent{\bf Caption-free Temporal Segment Splicing.} Evaluates MLLMs' ability to reconstruct shuffled video sequences using visual cues alone. We curated videos with inherent temporal logic (e.g., plant growth), segmented them into randomized clips, and tasked models with reordering via temporal reasoning.

\noindent{\bf Daily Life Action Evaluation.} Assesses MLLMs' ability to comparatively evaluate action quality in similar tasks. Using EPIC-Skills \cite{doughty2018s} videos with quality annotations, we reorganize clips into subtask pairs. Models identify the clip demonstrating superior execution quality through pairwise comparison.

\noindent{\bf Professional Action Evaluation.} Assesses MLLMs' capability for pairwise quality comparison of professional actions. Using AQA-7 \cite{parmar2019action}, we reorganize clips into pairs with significant referee-scored quality differences, requiring binary selection of the superior-quality video.

\noindent{\bf Weather Condition Evaluation.} Assesses MLLMs' ability to classify weather intensity through temporal dynamics and physical cues. We curated "rainy", "snowy", and "windy" videos at three intensity levels. Models perform comparative analysis on triplets per weather type to identify specified conditions.

\noindent{\bf Video Clarity Evaluation.} Assesses MLLMs' capability to evaluate perceptual video quality incorporating temporal artifacts from encoding/bitrates. Using CVQAD \cite{antsiferova2022video}, we curate triplet sequences with shuffled quality levels (high/medium/low) per content type. Models analyze temporal degradation patterns to restore original quality rankings.

\noindent{\bf Video Brightness Evaluation.} Assesses MLLMs' capability to evaluate scene illumination conditions. Using SDSD \cite{wang2021seeing} with controlled lighting variations, we generate triplet queries (1 dark + 2 bright or 2 dark + 1 bright) per trial. Models identify videos matching specified lighting conditions through multiple-choice selection.

\noindent{\bf Video Forensic Detection.} Evaluates MLLMs' ability to distinguish authentic versus AI-generated videos. Real videos exhibiting physical plausibility are paired with synthetic counterparts: captions generated via GPT-4o \cite{hurst2024gpt} drive video synthesis using Kling \cite{kling} and Ying \cite{ying}. Models identify authentic videos by detecting spatiotemporal inconsistencies through physical law reasoning.

\noindent{\bf Multiview Visual Perception Pairing.} Assesses MLLMs' capability for viewpoint-invariant object matching. Using paired clips from \cite{yonetani2016recognizing}, we present a reference video alongside three candidates (one correct match, two distractors). Models identify matching clips by analyzing cross-view correspondence through global object feature detection.

\noindent{\bf Common sense judgment of physical laws.} Probes MLLMs' understanding of physical causality in temporal dynamics. We curated natural phenomena videos (e.g., glacier melting, sand flow) with physically implausible reversed versions. Models discern plausible temporal logic from triplets containing one normal sequence and two inverted clips.

\noindent{\bf Gait Recognition Matching.} Evaluates MLLMs' cross-condition subject matching using CASIA Gait Database \cite{yu2006framework}. Given a reference gait sequence, models identify the same subject in candidate videos under varying conditions. Subjects with similar visual characteristics are deliberately paired to minimize confounding factors.

\noindent{\bf Cinematographic Style Matching.} Evaluates MLLMs' recognition of temporal editing techniques beyond static frames. We curated styles with distinctive temporal signatures (e.g., slow motion, time-lapse, frame extraction). Given a reference video and three candidates, models identify the optimal stylistic match through temporal pattern analysis.

\noindent{\bf Olympic Sports Matching.} Evaluates MLLMs' fine-grained motion discrimination within similar action categories. We curated sports sequences with high kinematic similarity (e.g., swimming strokes: butterfly, freestyle, breaststroke, backstroke), which static frames alone cannot reliably distinguish. Given a reference video and two same-category candidates, models discern subtle temporal variations.

\noindent{\bf Dance Style Matching.} Evaluates MLLMs' temporal pattern recognition across dance styles. We curated diverse styles with distinctive motion signatures (e.g., street, Latin, ballet), challenging to assess via static frames. Given a reference video and three candidates, models identify optimal stylistic matches through temporal dynamics analysis.

\section{Experiments}

\begin{table*}[!ht]
    \centering
    \caption{\textbf{Results of different models on MVPBench.} The best performance in each task is \textbf{in-bold}. Notably, the table presents normalized proficiency scores $\score_{mt}$ for each model rather than observed accuracy $\accuracy_{mt}$. Consequently, negative scores may occur, indicating model performance below random expectation levels for specific tasks. The "Overall" metric denotes a weighted average across models, with weights proportional to the information gain potential of each respective task.}
    \renewcommand{\arraystretch}{0.85}
    \resizebox{\textwidth}{!}{%
    \begin{tabular}{lcccccccc}
    \toprule
    \multicolumn{1}{l|}{Model} & Tem w/ Cap & Tem w/o Cap & Daily-Action & Pro-Action & Weather & Clarity & Brightness & Forensic \\
    \midrule
    \multicolumn{1}{l|}{\textit{Human}} & 95.16 & 80.01 & 80.96 & 85.72 & 83.34 & 90.00 & 95.71 & 97.00 \\
    \midrule
    \multicolumn{9}{c}{\textbf{\textit{Closed-source Multimodal LLMs}}} \\
    \midrule
    \multicolumn{1}{l|}{Gemini-2.5-Flash \cite{comanici2025gemini}} & \textbf{30.11} & \textbf{11.54} & 43.08 & 28.74 & \textbf{18.79} & 35.45 & 13.00 & 40.00 \\
    \multicolumn{1}{l|}{GPT-4o \cite{hurst2024gpt}} & 28.47 & 10.21 & \textbf{61.90} & \textbf{37.14} & 12.51 & 40.00 & 10.00 & \textbf{52.00} \\    
    \midrule
    \multicolumn{9}{c}{\textbf{\textit{Open-source Multimodal LLMs}}} \\
    \midrule
    \multicolumn{1}{l|}{MiniCPM-V 4.5 \cite{yu2025minicpm}} & 4.23 & -9.03 & -9.52 & -2.86 & -15.37 & -10.01 & 12.69 & 31.00 \\
    \multicolumn{1}{l|}{LLaVA-NeXT-Video-7B \cite{zhang2024video}} & -7.90 & -16.06 & 4.76 & 5.72 & -7.14 & -10.01 & 8.22 & -1.99 \\
    \multicolumn{1}{l|}{Qwen2.5-VL-7B \cite{bai2025qwen2}} & 9.07 & -5.12 & 19.04 & 5.72 & -16.66 & 26.28 & -13.33 & 22.00 \\
    \multicolumn{1}{l|}{Qwen2.5-VL-72B \cite{bai2025qwen2}} & 16.35 & 1.45 & 33.34 & 14.28 & -8.32 & 34.85 & 0.00 & 34.00 \\
    \multicolumn{1}{l|}{InternVL3-8B \cite{zhu2025internvl3}} & 9.07 & -11.92 & 23.80 & 5.72 & -9.99 & -13.33 & 13.00 & 10.00 \\
    \multicolumn{1}{l|}{InternVL3-38B \cite{zhu2025internvl3}} & 10.28 & -7.30 & 38.10 & 5.72 & -16.66 & 0.00 & 7.86 & 25.00 \\
    \multicolumn{1}{l|}{InternVL3-78B \cite{zhu2025internvl3}} & 12.71 & -2.92 & 38.10 & 14.28 & -3.33 & 6.66 & 20.71 & 34.00 \\
    \multicolumn{1}{l|}{Internlm-xcomposer2-7B \cite{dong2024internlm}} & 1.80 & -12.55 & -28.58 & -14.28 & -26.92 & 0.00 & 28.36 & 4.00 \\
    \multicolumn{1}{l|}{Internlm-xcomposer2.5-7B \cite{zhang2024internlm}} & 1.80 & -11.01 & -14.28 & -8.58 & -23.07 & 10.00 & \textbf{37.32} & 10.00 \\
    \multicolumn{1}{l|}{mPLUG-Owl3-7B \cite{ye2024mplug}} & 11.50 & -4.46 & -19.04 & -11.42 & -19.99 & 6.66 & 29.29 & 13.00 \\
    \multicolumn{1}{l|}{Emu3-Chat \cite{wang2024emu3}} & -7.90 & -16.06 & -9.52 & -2.86 & -3.84 & -20.00 & -17.85 & -7.99 \\
    \multicolumn{1}{l|}{LLaVA-OneVision-Qwen2-7B \cite{li2024llava}} & 13.92 & -0.73 & 33.34 & 8.58 & -12.49 & 31.43 & -6.66 & 31.00 \\
    \multicolumn{1}{l|}{LLaVA-OneVision-Qwen2-72B \cite{li2024llava}} & 22.41 & 8.02 & \textbf{61.90} & 25.72 & 0.00 & \textbf{41.71} & 3.34 & 40.00 \\
    \midrule
    \midrule
    \multicolumn{1}{l|}{Model} & Multi-view & Common-Sense & Gait & Cinematographic & Olympic & Dance & \multicolumn{2}{||c}{\textbf{Overall}} \\
    \midrule
    \multicolumn{1}{l|}{\textit{Human}} & 82.31 & 88.00 & 92.50 & 88.00 & 85.00 & 96.94 & \multicolumn{2}{||c}{88.89} \\
    \midrule
    \multicolumn{9}{c}{\textbf{\textit{Closed-source Multimodal LLMs}}} \\
    \midrule
    \multicolumn{1}{l|}{Gemini-2.5-Flash \cite{comanici2025gemini}} & 58.50 & \textbf{38.17} & 24.24 & \textbf{37.00} & 16.00 & \textbf{19.38} & \multicolumn{2}{||c}{29.33} \\
    \multicolumn{1}{l|}{GPT-4o \cite{hurst2024gpt}} & 65.01 & 32.65 & 31.00 & 32.50 & 12.00 & \textbf{19.38} & \multicolumn{2}{||c}{\textbf{31.10}} \\ 
    \midrule
    \multicolumn{9}{c}{\textbf{\textit{Open-source Multimodal LLMs}}} \\
    \midrule
    \multicolumn{1}{l|}{MiniCPM-V 4.5 \cite{yu2025minicpm}} & 70.00 & 8.17 & -3.49 & -6.49 & \textbf{37.14} & 11.88 & \multicolumn{2}{||c}{7.66} \\
    \multicolumn{1}{l|}{LLaVA-NeXT-Video-7B \cite{zhang2024video}} & 2.50 & 0.51 & -3.49 & -6.49 & -5.56 & -12.49 & \multicolumn{2}{||c}{4.30} \\
    \multicolumn{1}{l|}{Qwen2.5-VL-7B \cite{bai2025qwen2}} & 55.00 & 11.23 & 24.24 & 19.00 & -4.00 & -1.24 & \multicolumn{2}{||c}{10.90} \\
    \multicolumn{1}{l|}{Qwen2.5-VL-72B \cite{bai2025qwen2}} & 67.50 & 21.94 & \textbf{31.83} & 23.50 & 4.00 & 6.25 & \multicolumn{2}{||c}{20.00} \\
    \multicolumn{1}{l|}{InternVL3-8B \cite{zhu2025internvl3}} & 15.00 & 8.17 & 10.00 & 10.00 & 11.12 & -10.62 & \multicolumn{2}{||c}{3.99} \\
    \multicolumn{1}{l|}{InternVL3-38B \cite{zhu2025internvl3}} & 20.01 & 12.76 & 14.50 & 17.50 & -20.00 & -4.99 & \multicolumn{2}{||c}{6.97} \\
    \multicolumn{1}{l|}{InternVL3-78B \cite{zhu2025internvl3}} & 35.01 & 18.88 & 17.50 & 23.50 & -12.00 & 0.63 & \multicolumn{2}{||c}{14.06} \\
    \multicolumn{1}{l|}{Internlm-xcomposer2-7B \cite{dong2024internlm}} & 0.00 & -4.08 & -3.49 & -6.49 & -12.00 & -10.62 & \multicolumn{2}{||c}{-5.22} \\
    \multicolumn{1}{l|}{Internlm-xcomposer2.5-7B \cite{zhang2024internlm}} & 12.51 & 11.23 & 13.00 & 8.50 & -12.00 & -3.12 & \multicolumn{2}{||c}{2.89} \\
    \multicolumn{1}{l|}{mPLUG-Owl3-7B \cite{ye2024mplug}} & 35.01 & -12.49 & 1.00 & 10.00 & -16.66 & 9.28 & \multicolumn{2}{||c}{3.41} \\
    \multicolumn{1}{l|}{Emu3-Chat \cite{wang2024emu3}} & -12.49 & -8.67 & -3.49 & -18.49 & -6.66 & -10.62 & \multicolumn{2}{||c}{-10.89} \\
    \multicolumn{1}{l|}{LLaVA-OneVision-Qwen2-7B \cite{li2024llava}} & \textbf{72.04} & 11.23 & 17.50 & 11.50 & 4.00 & -1.24 & \multicolumn{2}{||c}{15.16} \\
    \multicolumn{1}{l|}{LLaVA-OneVision-Qwen2-72B \cite{li2024llava}} & \textbf{72.04} & 23.47 & 25.00 & 20.50 & 16.00 & 10.00 & \multicolumn{2}{||c}{25.79} \\
    \bottomrule
    \end{tabular}%
    }
    \label{tab:3}
    \end{table*}

\subsection{Experimental Setup}

\noindent{\bf Evaluation setup:} To mitigate bias, we standardize prompts to explicitly enumerate video roles as noted below. This design ensures equitable evaluation across heterogeneous model capabilities:
{\small
\begin{equation}
Prompt+Video1:\textless video\textgreater+\ldots+VideoN:\textless video\textgreater+Prompt \nonumber
\end{equation}
}

In the prompt template, \textless video\textgreater serves as a placeholder for videos, with the model reserving these positions for subsequent input video visual tokens. To differentiate between various input videos, distinct video encodings are added before the placeholder. While the required form of model output varies across different tasks, we consistently require the model to produce strictly numerical outputs. This facilitates a straightforward horizontal comparison of the results.

We employ a set of predefined rules along with GPT-4-turbo \cite{openai2024gpt} to extract the selected answer from the model's output. We perform manual verification on 15\% of outputs with 98\% MLLM-Human Agreement, validating the robustness of our automated extraction pipeline.

\noindent{\bf Human baseline:}  We adopted a stratified sampling approach to establish reliable human performance benchmarks for MVPBench, with 52 ± 3 questions per evaluator(20 evaluators in total). We calculated the inter-evaluator agreement (89.7\%) to verify the rationality of the human baseline. Each evaluator received 10\% randomly interspersed duplicate questions (104 total) from other evaluators' sets. For each duplicate pair $(q_i, q_j)$ answered by evaluators $(E_m, E_n)$, agreement was scored as:
\begin{equation}
    \text{Agreement} = \frac{\sum_{k=1}^{N} \mathbb{I}(A_{m,k} = A_{n,k})}{N} \times 100\%,
    \label{eq:agreement}
\end{equation}
where $\mathbb{I}(\cdot)$ denotes the indicator function, $N$ is the number of duplicate questions for consistency, and $A_{m,k}$ represents evaluator $m$'s response to the $k$-th question.


\noindent{\bf Standardized Performance Assessment:} To ensure a statistically rigorous framework for comparing model capabilities across heterogeneous subtasks with divergent chance performance levels, we adopt a standardized evaluation protocol grounded in information-theoretic principles. Let $\chance_t$ denote the \emph{chance performance} for subtask $t$, defined as $\chance_t = 1/k_t$, where $k_t$ represents the number of response options. For a model's observed accuracy $\accuracy_{mt}$ on subtask $t$, we compute its \emph{normalized proficiency score} $\score_{mt}$ as:
\begin{equation}
\score_{mt} = \frac{\accuracy_{mt} - \chance_t}{1 - \chance_t}
\end{equation}

Aggregate capability scores $\bar{\mathscr{S}}_m$ for model $m$ across $N$ subtasks are derived through weighted averaging of $\score_{mt}$ across tasks, with weights proportional to the task's \emph{information gain potential} ($1 - \chance_t$), thereby prioritizing tasks with higher discriminative power.
\begin{equation}
\bar{\mathscr{S}}_m = \frac{\sum_{t=1}^N w_t \cdot \mathscr{S}_{mt}}{\sum_{t=1}^N w_t}, \quad \text{where} \quad w_t = 1 - \mathfrak{C}_t
\end{equation}

\subsection{Analysis}

\begin{itemize}[leftmargin=*, itemsep=0pt, parsep=0pt, topsep=0pt] 
    \item \textbf{Temporal Splicing (w/ \& w/o Caption):} Lowest accuracy (10.39\%; -4.40\%), indicating core temporal reasoning limitations. Marginally better caption-guided performance suggests text partially mitigates challenges, but models still struggle with multi-video sequencing. Strong inter-task correlation confirms shared dependence on seriality—poorly handled by current architectures.
    \item \textbf{Action Evaluation (Daily/Professional):} Moderate accuracy (Daily: 18.43\%; Professional: 7.44\%) with positive cross-task correlation. The performance gap reveals scaling difficulty with temporal specificity, confirming context-dependent action comprehension. Inclusion of both subtasks prevents bias toward generic/domain-specific understanding. 
    \item \textbf{Video Quality Assessment (Clarity/Brightness/Forensic):} Divergent trends: Clarity (11.98\%) and Brightness (9.73\%) rely on low-level features, while Forensic (22.40\%) requires semantic reasoning. Weak inter-task correlation indicates quality assessment comprises distinct subskills—from pixel analysis to physical plausibility.
    \item \textbf{Logic Inference (Multiview/Common-Sense):} Highest accuracy in Multiview Pairing (37.84\%, aligning with static matching), versus Common-Sense Judgment (11.54\%, requiring temporal causality). This dichotomy validates MVPBench's coverage of static/dynamic reasoning.
    \item \textbf{Similarity Matching (Gait/Style/Sports/Dance):} Highly variable performance (Gait: 13.06\%; Cinematographic: 11.70\%; Olympic: 9.48\%; Dance: 1.46\%). Weak correlations confirm these probe distinct temporal dimensions: motion patterns (Gait/Dance) versus stylistic techniques (Cinematographic).
\end{itemize}

Collectively, our experimental findings suggest that current multimodal models exhibit significant limitations in processing and integrating multiple video inputs. As illustrated in Tab. \ref{tab:3}, some models \cite{dong2024internlm, wang2024emu3} not only underperform in complex multi-video tasks but also fail to surpass even random-chance baselines. We also found that current models exhibit systematic biases when processing multi-video inputs (e.g., \textit{the model tends to respond 1, 2, 3, ... in the splicing task, and chooses the first video more frequently than other videos in the matching task}). These biases suggest that rather than genuinely analyzing and integrating visual information, the models rely on superficial heuristics or textual priors, leading to suboptimal and inconsistent reasoning. In terms of the model's positional bias toward the first video in candidate rankings, we have performed additional analyses, the results are shown in Tab. \ref{tab:5}. When ground-truth answers are uniformly distributed, preferentially selecting the first option guarantees performance matching random chance. By intentionally reducing first-option frequency in our ground-truth, we avoid this situation—explaining why models frequently underperform random baselines in our results.

\begin{table*}
\centering
\caption{\textbf{The result of first-option frequency evaluated on Qwen2.5-VL-72B and its random expectation.} The numbers represent MLLM-Human Agreement. The model selects the first video option at a frequency 1.5–2 times the random expectation threshold, empirically validating the presence of Positional Bias in our experimental framework.}
\renewcommand{\arraystretch}{0.85}
\resizebox{\textwidth}{!}{
\begin{tabular}{l|cccccc}
\toprule
\textbf{Baseline} & \textbf{Tem w/ Cap}$\uparrow$ & \textbf{Tem w/o Cap}$\uparrow$ & \textbf{Gait}$\uparrow$ & \textbf{Cinematographic}$\uparrow$ & \textbf{Olympic}$\uparrow$ & \textbf{Dance}$\uparrow$ \\
\midrule
Model & 21.13\% & 28.30\% & 49.00\% & 68.00\% & 78.00\% & 46.25\% \\
Random Expectation & 10.56\% & 16.67\% & 33.33\% & 33.33\% & 50.00\% & 33.33\% \\
\bottomrule
\end{tabular}
}
\label{tab:5}
\end{table*}


\subsection{Ablation Study}

To clarify the essential distinctions between single-video benchmarks, we conducted a comparative experiment to determine whether adopting the following alternative approach would result in a significant performance gap compared to the multi-video input method presented in our article: 
\begin{enumerate}
    \item First, compute task-specific metrics for individual videos, then perform a comparative analysis of these metrics.
    \item First, sequentially process individual videos to generate descriptive captions, then provide the aggregated captions to the model for comprehensive question answering.
\end{enumerate}

As illustrated in Tab. \ref{tab:6}, the model exhibits a significant performance degradation across the single-video style benchmark experiments on all evaluation tasks relative to our multi-video methodology. Our experiments demonstrate that single-video style evaluation methodologies invariably produce discrepancies between assessed performance and the model's core competencies.

To better showcase our experimental setup, we selected two diving videos exhibiting significant divergence in human-assigned quality ratings as examples. We processed both videos through the model and evaluated them using the two aforementioned methodologies. The experimental results and potential explanations for the inconsistency in performance are as below:
    
\noindent{\bf{Single video input + Task-specific metrics comparison.}}

\noindent{\textit{Video A (Human score: 98.4) → Caption: "Performed well" → Score: 7.0}}

\noindent{\textit{Video B (Human score: 29.7) → Caption: "Lack of Liquidity", "Imperfect" → Score: 7.8}}

\begin{itemize}[leftmargin=*, itemsep=0pt, parsep=0pt, topsep=0pt] 
    \item Limited domain expertise makes MLLMs struggle to assign precise numeric ratings, whereas comparing multiple videos side by side lets the model make relative, ranking-like judgments—proven in InstructGPT that ranking-based fine‑tuning outperform absolute scoring.
    \item Lack of unified scoring baseline. Each video is scored independently, so a “7” for Video A may represent very strong performance in that context, while a “7.8” for Video B may actually signal weaker performance. Absence of multi-video's visual information and comparability prevents direct judgement of these scores.
\end{itemize}

\noindent{\bf{Single video input + Caption comparison.}}

\noindent{\textit{Video A (Human score: 98.4) → Caption: "Forward 3½ Somersaults Pike (DD 3.2)", “Exceptional body control and stability”, “Rushed takeoff and imperfect pike shape” → Score: 7.8}}

\noindent{\textit{Video B (Human Score: 29.7) → Caption: "Forward 2½ Somersaults Pike (DD 2.8)", "Minor crown splash", “Good rhythm and slight reduction in fluidity” → Score: 8.1}}

\begin{itemize}[leftmargin=*, itemsep=0pt, parsep=0pt, topsep=0pt] 
    \item Models cannot reliably distill discriminative visual details into text (e.g., The extent of the splash when entering the water is not fully demonstrated).
    \item Subtle differences (e.g., degree of body tilt) become indistinguishable in text formats. Consequently, the model fails to discriminate salient actions when processing undifferentiated textual descriptions.
    \item Lack of visual information for quantitative scoring. 
\end{itemize}

\begin{table}[htbp]
\centering
\caption{\textbf{Ablation study across the single-video style benchmark experiments on all evaluation tasks.} All experimental values represent the mean MLLM-Human Agreement scores \cite{gu2024survey}, averaged across models with native multi-video processing capabilities (Gemini-2.5-Flash, GPT-4o, and LLaVA-OneVision-Qwen2).}
\label{tab:6}

\begin{subtable}{\columnwidth}
\centering
\caption{The results of single-video input followed by scores comparison and direct multi-video comparison.}
\label{tab:ablation1}
\resizebox{\linewidth}{!}{
\begin{tabular}{@{}p{2.5cm}ccccc@{}}
\toprule
\textbf{Subtasks} & VQA & VLI & TSS & CA & SVP\\
\midrule
Random & 33.33\% & 33.33\% & 11.41\% & 45.22\% & 35.86\% \\
Single-video & 39.10\% & 37.97\% & 11.41\% & 51.59\% & 39.70\% \\
\cmidrule(r){1-6}
\textbf{Multi-video} & \textbf{58.97\%} & \textbf{65.19\%} & \textbf{26.76\%} & \textbf{61.15\%} & \textbf{60.61\%} \\
\bottomrule
\end{tabular}
}
\end{subtable}

\vspace{1em}

\begin{subtable}{\columnwidth}
\centering
\caption{The results of generating captions followed by textual comparison and direct multi-video comparison.}
\label{tab:ablation2}
\resizebox{\linewidth}{!}{
\begin{tabular}{@{}p{2.5cm}ccccc@{}}
\toprule
\textbf{Subtasks} & VQA & VLI & TSS & CA & SVP\\
\midrule
Random & 33.33\% & 33.33\% & 11.41\% & 45.22\% & 35.86\% \\
Caption Comparison & 34.62\% & 53.80\% & 14.08\% & 50.96\% & 46.36\% \\
\cmidrule(r){1-6}
\textbf{Video Comparison} & \textbf{58.97\%} & \textbf{65.19\%} & \textbf{26.76\%} & \textbf{61.15\%} & \textbf{60.61\%} \\
\bottomrule
\end{tabular}
}
\end{subtable}
\end{table}

\section{Conclusion}

We introduce MVPBench, a benchmark designed to evaluate the perceptual understanding capabilities of MLLMs when dealing with multiple video inputs. Our experimental results indicate that these multi-video benchmarks present significant challenges to a wide range of current advanced MLLMs. Through our benchmark, we aim to stimulate further research in this area, encouraging exploration into enhancing the model's ability to comprehend multi-video inputs. MVPBench is positioned to serve as a comprehensive and integrated testing platform for evaluating related models, thereby supporting advancements in this field.

\section{Details of Prompt Design}
\label{sec:B}
The specific prompt design for each subtask is detailed in Table \ref{tab:5}, Table \ref{tab:6} and Table \ref{tab:7}. We have standardized the structure of each question prompt based on the task type. Each prompt comprises the task description, video encoding explanation, task action goal, model output requirements, input video information, and any additional input information. This modular design ensures clarity in the model's target requirements for each subtask and minimizes the influence of extraneous factors on the model's ability to process multiple video inputs.

In the prompt template, \textless video\textgreater serves as a placeholder for videos, with the model reserving these positions for subsequent input video visual tokens. To differentiate between various input videos, distinct video encodings are added before the placeholder. For matching tasks, due to the design requirements, the prompt template includes both the video encoding and placeholder of the reference video, enabling the model to better achieve the task objectives. While the required form of model output varies across different tasks, we consistently require the model to produce strictly numerical outputs. This facilitates a straightforward horizontal comparison of the results.

\begin{table*}[ht]
    \centering
    \begin{tabular}{m{0.15\textwidth}|>{\columncolor{gray!20}}m{0.75\textwidth}}
        \hline\hline
        \multicolumn{1}{c|}{\textbf{Subtask}} & \textbf{Prompt example} \\ \hline\hline
        \centering Caption-guided temporal segment splicing & \textit{You are tasked with sorting video clips that correspond to steps in a recipe. Each video is numbered sequentially based on the order of input, such as 1, 2, 3, and so on. The videos are currently in a shuffled order, and your goal is to arrange them according to the correct sequence described in the recipe. The final output should be a sequence of numbers representing the correct order of the videos without any additional information. Here is the information provided:}
        
        \textit{Video 1: \textless video\textgreater Video 2: \textless video\textgreater Video 3: \textless video\textgreater} 
        
        \textit{Recipe Description:
        Step 0: Put foil over the pan.
        Step 1: Put kalbi strips in the pan.
        Step 2: Put in the broiler.}
        
        \textit{Please determine and provide the correct sequence of video numbers.} \\ \hline
        \centering Caption-free temporal segment splicing & \textit{You are tasked with sorting video clips based on their content or logical sequence. Each video is numbered sequentially based on the order of input, such as 1, 2, 3, and so on. The videos are currently in a shuffled order, and your goal is to determine the correct order of these video clips. The final output should be a sequence of numbers representing the correct order of the videos without any additional information. Here are the video clips:}

        \textit{Video 1: \textless video\textgreater Video 2: \textless video\textgreater Video 3: \textless video\textgreater} 

        \textit{Please determine and provide the correct sequence of video numbers.}
        \\ \hline
        \centering Daily life action evaluation & \textit{Please watch the two video clips provided, labeled as Video 1 and Video 2. Each clip contains similar content and tasks being performed. Your task is to compare these clips and evaluate which one demonstrates better task and action completion. After reviewing both videos, select the number corresponding to the video that performed the task more effectively. Your response should be either the number 1 or 2, without any additional commentary or explanation. Here are the video clips:}
        
        \textit{Video 1: \textless video\textgreater Video 2: \textless video\textgreater} \\ \hline
        \centering Professional action evaluation & \textit{Please watch the two video clips provided, labeled as Video 1 and Video 2. Each clip contains similar content and tasks being performed. Your task is to compare these clips and evaluate which one demonstrates better task and action completion. After reviewing both videos, select the number corresponding to the video that performed the task more effectively. Your response should be either the number 1 or 2, without any additional commentary or explanation. Here are the video clips:}
        
        \textit{Video 1: \textless video\textgreater Video 2: \textless video\textgreater} \\ \hline
        \centering Weather condition evaluation & \textit{Please evaluate the three provided video clips, labeled as Video 1, Video 2, and Video 3. Each clip depicts the same type of weather, but with varying intensity levels: severe, moderate, and mild. Your task is to examine the weather conditions in each video to identify which one exhibits the most severe weather. After assessing the videos, provide a single digit corresponding to the video number that shows the most severe weather conditions. Your response should be concise, consisting solely of the digit without any additional commentary or explanation. Here are the video clips:}
        
        \textit{Video 1: \textless video\textgreater Video 2: \textless video\textgreater Video 3: \textless video\textgreater} \\ \hline\hline   
    \end{tabular}
    \caption{Prompt examples of \textbf{Temporal Segment Splicing} tasks and \textbf{Content Assessment} tasks.}
    \label{tab:5}
\end{table*}

\begin{table*}[ht]
    \centering
    \begin{tabular}{m{0.15\textwidth}|>{\columncolor{gray!20}}m{0.75\textwidth}}
        \hline\hline
        \multicolumn{1}{c|}{\textbf{Subtask}} & \textbf{Prompt example} \\ \hline\hline
        \centering Video clarity evaluation & \textit{Please evaluate the three provided video clips, labeled as Video 1, Video 2, and Video 3, focusing on their video quality, specifically in terms of clarity and smoothness. Your task is to rank these videos from highest to lowest quality based solely on these criteria. Indicate your ranking by listing only the corresponding numbers of each video in the order of quality, such as "3, 2, 1" if Video 3 is the highest quality, followed by Video 2, and then Video 1. Ensure your response includes only the ranking numbers without any additional explanation or commentary. Here are the video clips:}
        
        \textit{Video 1: \textless video\textgreater Video 2: \textless video\textgreater Video 3: \textless video\textgreater} \\ \hline
        \centering Video brightness evaluation & \textit{Please analyze the brightness levels of the three provided video clips, labeled as Video 1, Video 2, and Video 3, each with distinct lighting and shooting conditions. Your task is to determine which video clip exhibits the lowest brightness and was filmed under the darkest conditions. After your assessment, provide your answer as a single digit representing the video number with the lowest brightness. Your response should be concise, consisting solely of the digit without any additional commentary or explanation. Here are the video clips:}
        
        \textit{Video 1: \textless video\textgreater Video 2: \textless video\textgreater Video 3: \textless video\textgreater} \\ \hline
        \centering Video forensic detection & \textit{Please evaluate the three provided video clips, labeled as Video 1, Video 2, and Video 3. Each clip contains similar content, but two are AI-generated, and one is authentic. Your task is to compare these videos and determine which one is real. After reviewing all three, respond with the number of the video you believe is genuine. Your response should be a single digit, representing the video number, without any additional commentary or explanation. Here are the video clips:} 

        \textit{Video 1: \textless video\textgreater Video 2: \textless video\textgreater Video 3: \textless video\textgreater} 
        \\ \hline
        \centering Multiview visual perception pairing & \textit{Please analyze the given reference video,}
        
        \textit{Ref Video: \textless video\textgreater}
        
        \textit{Compare it with the three candidate videos provided:} 
        
        \textit{Video 1: \textless video\textgreater Video 2: \textless video\textgreater Video 3: \textless video\textgreater} 
        
        \textit{The task is to identify the candidate video that corresponds to the reference video. The reference video is presented from a first-person perspective, while the correct candidate video is shown from a second-person perspective of the same scene. The other two candidate videos are considered distractors and do not match the reference video. Please examine each video closely and identify the candidate video that accurately matches the reference video. Provide your answer solely as a single digit that corresponds to the number of the correct candidate video, without any additional explanation or information.}   
        \\ \hline
        \centering Common sense judgment of physical laws & \textit{Please evaluate the three provided video clips, labeled as Video 1, Video 2, and Video 3. Among these clips, two contain elements that clearly violate natural physical laws, while one depicts a scenario consistent with typical physical principles. Your task is to identify the video that aligns with normal physical laws. After reviewing all three, respond with the number of the video indicating the number of the video that follows the laws of physics. Your response should  be a single digit, representing the video number, without any additional commentary or explanation. Here are the video clips:} 

        \textit{Video 1: \textless video\textgreater Video 2: \textless video\textgreater Video 3: \textless video\textgreater} 
        \\ \hline\hline
    \end{tabular}
    \caption{Prompt examples of \textbf{Video Quality Assessment} tasks and \textbf{Video Logic Inference} tasks.}
    \label{tab:6}
\end{table*}

\begin{table*}[ht]
    \centering
    \begin{tabular}{m{0.15\textwidth}|>{\columncolor{gray!20}}m{0.75\textwidth}}
        \hline\hline
        \multicolumn{1}{c|}{\textbf{Subtask}} & \textbf{Prompt example} \\ \hline\hline
        \centering Gait recognition matching & \textit{Please analyze the given reference video,}
        
        \textit{Ref Video: \textless video\textgreater}
        
        \textit{Compare it with the three candidate videos provided:} 
        
        \textit{Video 1: \textless video\textgreater Video 2: \textless video\textgreater Video 3: \textless video\textgreater} 
        
        \textit{The task is to identify which candidate video features the same individual as the reference video, based on their gait, which should be identical. The other two candidate videos are distractors, featuring different individuals with different gaits. Please examine each video closely and identify the candidate video that accurately matches the reference video. Provide your answer solely as a single digit that corresponds to the number of the correct candidate video, without any additional explanation or information.}
        \\ \hline
        \centering Shooting style matching & \textit{Please analyze the given reference video,}
        
        \textit{Ref Video: \textless video\textgreater}
        
        \textit{Compare it with the three candidate videos provided:} 
        
        \textit{Video 1: \textless video\textgreater Video 2: \textless video\textgreater Video 3: \textless video\textgreater} 
        
        \textit{Your task is to identify which candidate video shares the same filming style as the reference video, focusing specifically on camera movement and video processing techniques. Among the three candidate videos, only one will match the reference video in these aspects, while the other two will not. Please examine each video closely and identify the candidate video that accurately matches the reference video. Provide your answer solely as a single digit that corresponds to the number of the correct candidate video, without any additional explanation or information.}
        \\ \hline
        \centering Olympic sports matching & \textit{Please analyze the given reference video,}
        
        \textit{Ref Video: \textless video\textgreater}
        
        \textit{Compare it with the three candidate videos provided:} 
        
        \textit{Video 1: \textless video\textgreater Video 2: \textless video\textgreater Video 3: \textless video\textgreater} 
        
        \textit{Your task is to determine which of the candidate videos depicts the same sports activity as the reference video. Note that one candidate video matches the reference video while the others serves as a distractor. Please examine each video closely and identify the candidate video that accurately matches the reference video. Provide your answer solely as a single digit that corresponds to the number of the correct candidate video, without any additional explanation or information.}
        \\ \hline
        \centering Dance style matching & \textit{Please analyze the given reference video,}
        
        \textit{Ref Video: \textless video\textgreater}
        
        \textit{Compare it with the three candidate videos provided:} 
        
        \textit{Video 1: \textless video\textgreater Video 2: \textless video\textgreater Video 3: \textless video\textgreater} 
        
        \textit{Your task is to determine which of the candidate videos features the same dance type as the reference video. Note that one candidate video matches the reference video while the others serves as a distractor. Please examine each video closely and identify the candidate video that accurately matches the reference video. Provide your answer solely as a single digit that corresponds to the number of the correct candidate video, without any additional explanation or information.}
        \\ \hline\hline
    \end{tabular}
    \caption{Prompt examples of \textbf{Similar Video Pairing} tasks.}
    \label{tab:7}
\end{table*}

\section{Evaluation Prompts}
\label{sec:C}
Given model outputs, we extract the choices using GPT-3.5. We provide GPT with the model prediction, and then request GPT to extract the precise video number (or combination thereof) that the model predicted as the correct answer. Here is the prompt we use:

\hrulefill

\textbf{Prompt:}

\textit{You are given a passage that contains descriptions and discussions about various videos, including video numbers. Your task is to extract the video number(s) that represent the correct answer based on the context of the entire passage. The passage may contain multiple video numbers used for descriptions, analyses, or as part of reasoning processes. Ignore any video numbers that are not the final answer. Provide only the video number(s) that directly answer the question or represent the correct choice, without any additional text or explanation.}

\textbf{Passage:}

[Insert Passage Here]

\hrulefill

\textbf{Example:}

\textbf{Passage:}

\emph{"The video with the most severe weather conditions is Video 3."}

\textbf{Extracted Information:}

\texttt{3}

\textbf{Passage:}

\emph{"The correct order of the video clips is: 1, 2, 3."}

\textbf{Extracted Information:}

\texttt{1, 2, 3}

\textbf{Passage:}

\emph{"Video 1: This video shows a natural scene... However, if you are looking for a single video that represents a more complex or visually striking example, Video 1 might be the most visually captivating due to its natural and dynamic ice arch formation in an arctic environment."}

\textbf{Extracted Information:}

\texttt{1}

\hrulefill

\textbf{Instructions:}

\begin{itemize}
    \item Read the entire passage carefully.
    \item Determine which video number(s) represent the final answer based on the context.
    \item Ignore any video numbers mentioned only for description or analysis purposes.
    \item Provide \textbf{only} the number(s) of the video(s) that are the correct answer, separated by commas if more than one.
    \item Do not include any additional text, explanations, or formatting.
\end{itemize}

\section{Subtask Setup}
\label{sec:D}
\noindent{\bf Weather condition evaluation.} 
\textit{Determining wind strength (the "windy" problem) requires evaluating the magnitude and frequency of positional changes in reference objects over a given time interval in the video, whereas a static image lacks sufficient information for such judgment. Similarly, identifying "rainy" or "snowy" conditions also necessitates temporal analysis, which is why we deliberately selected these scenarios to evaluate the model's ability to process time-domain information.}

\noindent{\bf Video clarity evaluation.} 
\textit{This task assesses the model's video quality assessment (VQA) capability. Unlike image quality assessment (IQA), video quality cannot be accurately inferred from a single frame due to dynamic factors like encoding methods and bitrates. Compared to no-reference VQA (NR-VQA), our method presents multiple quality-variant videos of the same content, enabling reference-based evaluation while reducing cross-model subjective bias.}

\noindent{\bf Video brightness evaluation.} 
\textit{This task employs a more lenient evaluation criterion for assessing models' utilization of temporal information, primarily emphasizing the model's capability to process and integrate \textbf{multiple video inputs}, rather than the perception and reasoning capabilities in the temporal dimension compared to the single frame input.}

\noindent{\bf Dance style matching.}
\textit{The objective of similar video pairing aligns with the video embedding problem, aiming to evaluate whether the model can effectively capture inter-video similarities. This capability has direct implications for the model's potential applications in video recommendation systems, content-based video retrieval, and copyright infringement detection.}

\section{Ablation Study Examples}
\label{sec:E}

To better showcase our experimental setup, we selected two diving videos exhibiting significant divergence in human-assigned quality ratings as examples. The higher-rated video featured an athlete performing dives with a higher degree of difficulty, demonstrating smoother overall water entry and minimal splash displacement. Conversely, the lower-rated video displayed simpler maneuvers and generated substantially greater splash upon water entry. We processed both videos through the model and evaluated them using the two aforementioned methodologies. The experimental results and potential explanations for the inconsistency in performance are as below:

\noindent{\bf{Single video input + Task-specific metrics comparison.}}

\noindent{\textit{Video A (Human score: 98.4) → Caption: "Performed well" → Score: 7.0}}

\noindent{\textit{Video B (Human score: 29.7) → Caption: "Lack of Liquidity", "Imperfect" → Score: 7.8}}

\begin{itemize}[leftmargin=*, itemsep=0pt, parsep=0pt, topsep=0pt] 
    \item Limited domain expertise makes MLLMs struggle to assign precise numeric ratings, whereas comparing multiple videos side by side lets the model make relative, ranking-like judgments—proven in InstructGPT that ranking-based fine‑tuning outperform absolute scoring.
    \item Lack of unified scoring baseline. Each video is scored independently, so a “7” for Video A may represent very strong performance in that context, while a “7.8” for Video B may actually signal weaker performance. Absence of multi-video's visual information and comparability prevents direct judgement of these scores.
\end{itemize}

\noindent{\bf{Single video input + Caption comparison.}}

\noindent{\textit{Video A (Human score: 98.4) → Caption: "Forward 3½ Somersaults Pike (DD 3.2)", “Exceptional body control and stability”, “Rushed takeoff and imperfect pike shape” → Score: 7.8}}

\noindent{\textit{Video B (Human Score: 29.7) → Caption: "Forward 2½ Somersaults Pike (DD 2.8)", "Minor crown splash", “Good rhythm and slight reduction in fluidity” → Score: 8.1}}

\begin{itemize}[leftmargin=*, itemsep=0pt, parsep=0pt, topsep=0pt] 
    \item Models cannot reliably distill discriminative visual details into text (e.g., The extent of the splash when entering the water is not fully demonstrated).
    \item Subtle differences (e.g., degree of body tilt) become indistinguishable in text formats. Consequently, the model fails to discriminate salient actions when processing undifferentiated textual descriptions.
    \item Lack of visual information for quantitative scoring. 
\end{itemize}

\section{Copyright}
\label{sec:F}
MVPBench is a research preview intended for non-commercial use only. Users must cite MVPBench in any derivative work and acknowledge all original data sources (Pexels, YouTube, etc.). For open-sourced datasets incorporated into MVPBench, we strictly follow their licenses and attribute them appropriately. By using MVPBench, you agree to abide by these guidelines and all applicable copyright regulations. Violations may result in revocation of access. The benchmark includes manually collected videos from multiple sources, each governed by corresponding usage terms.

\section{More Visualization}
\label{sec:G}
\begin{figure*}[ht]
    \centering
    \includegraphics[width=\textwidth]{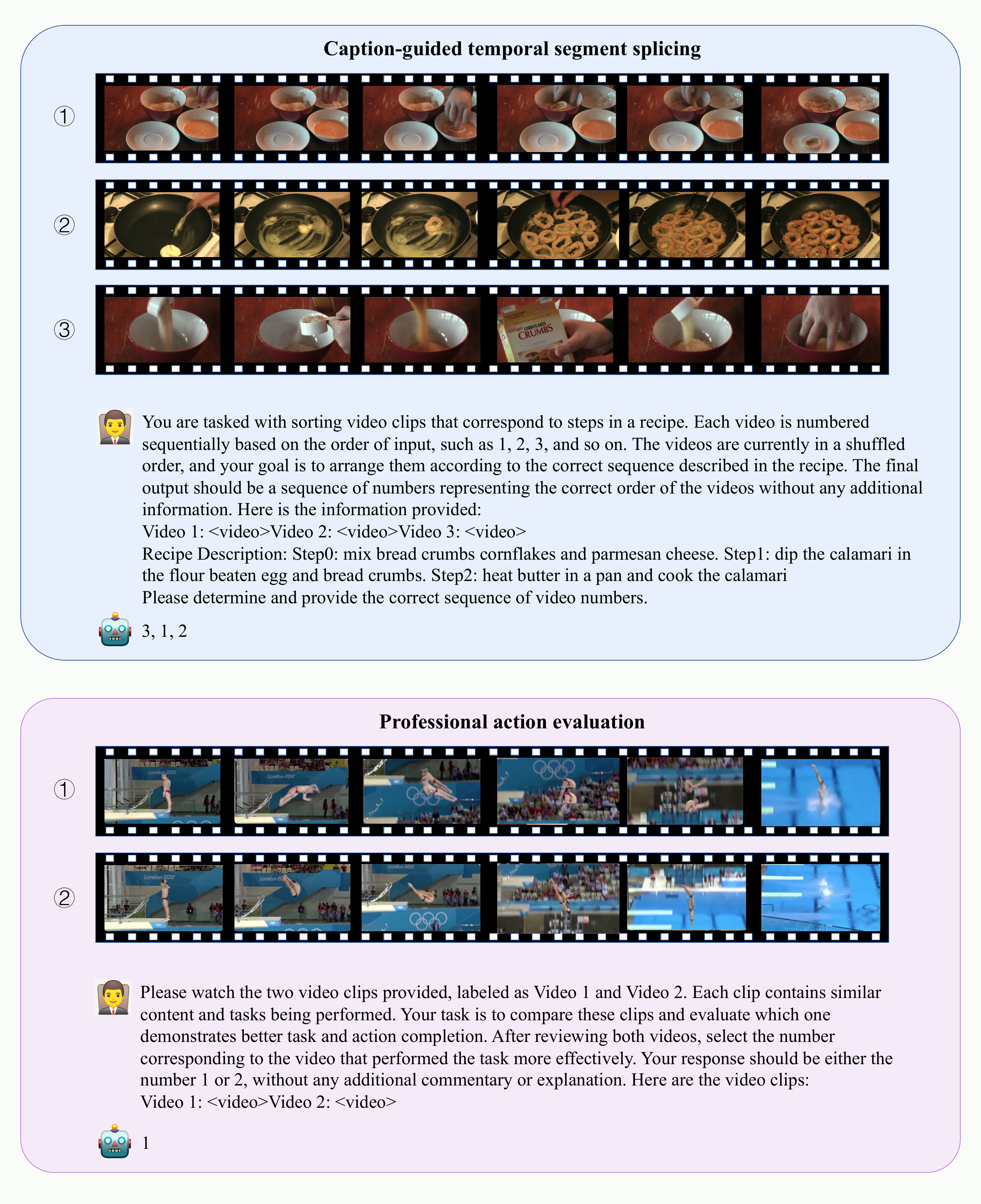}\vspace{-0.3cm} 
    \caption{\textbf{More visualization of data in MVPBench.} Examples of \textbf{Temporal Segment Splicing} tasks and \textbf{Content Assessment} tasks.}
    \label{fig:5}
\end{figure*}

\begin{figure*}[ht]
    \centering
    \includegraphics[width=\textwidth]{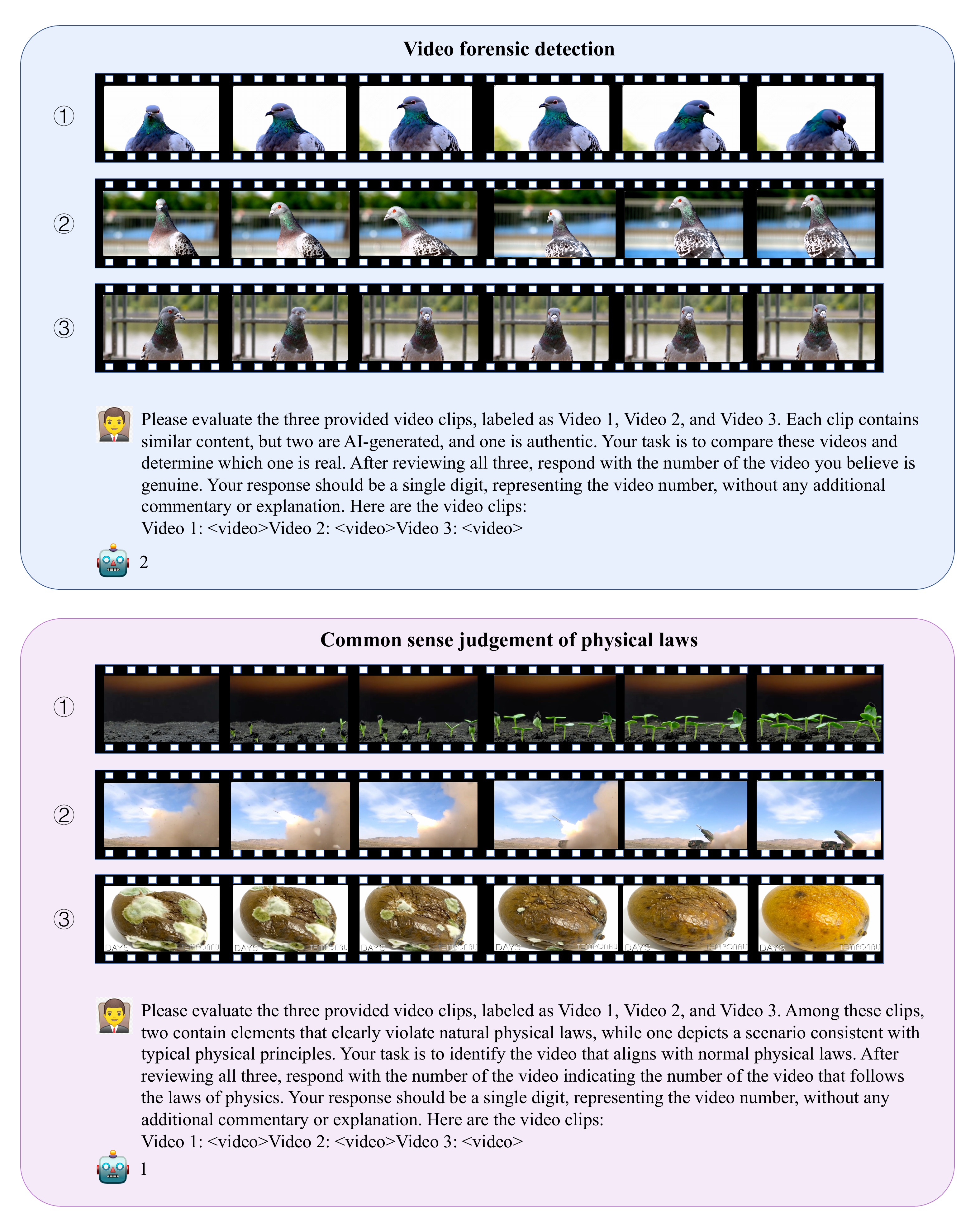}\vspace{-0.3cm} 
    \caption{\textbf{More visualization of data in MVPBench.} Examples of \textbf{Video Quality Assessment} tasks and \textbf{Video Logic Inference} tasks.}
    \label{fig:6}
\end{figure*}

\begin{figure*}[ht]
    \centering
    \includegraphics[width=\textwidth]{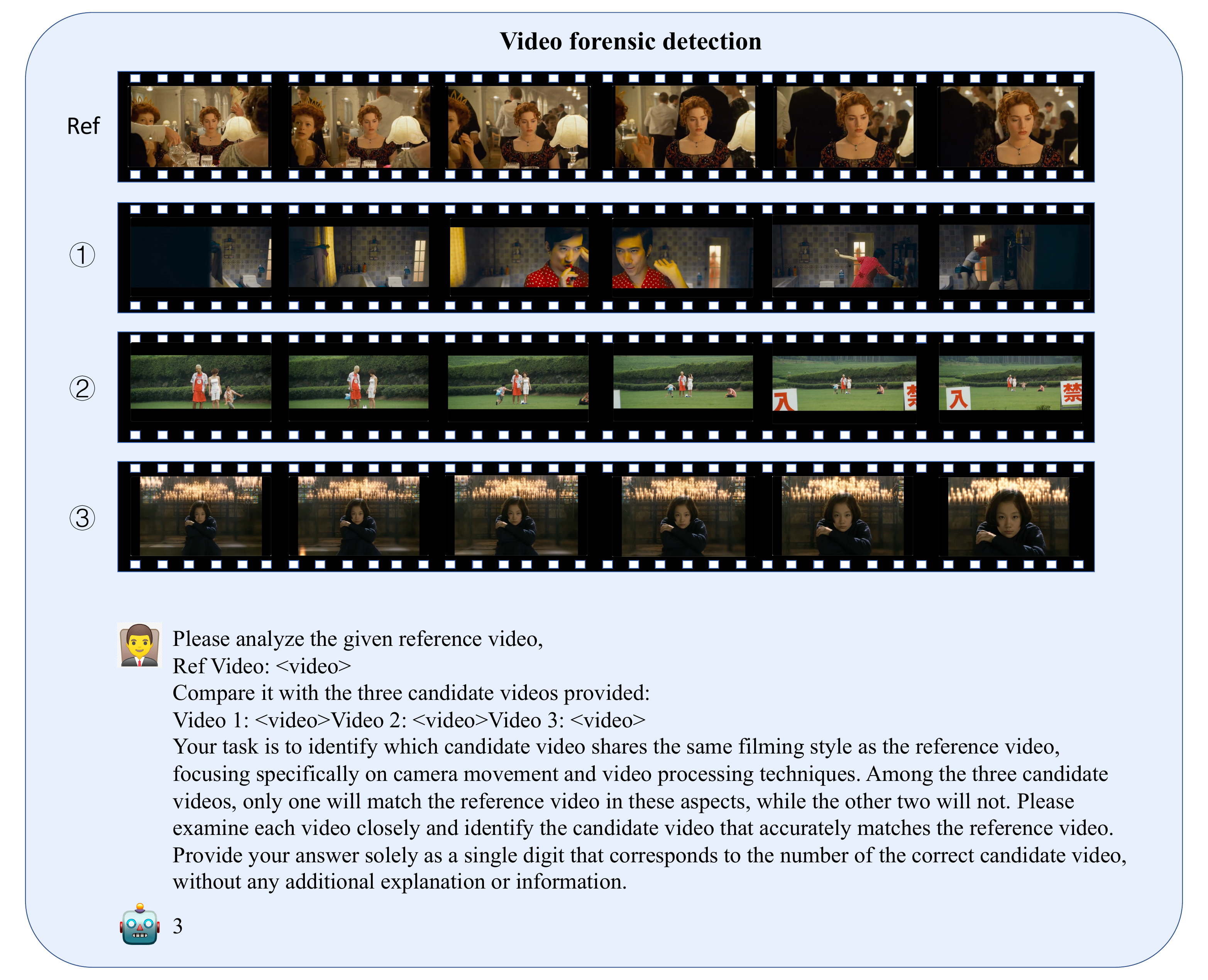}\vspace{-0.3cm} 
    \caption{\textbf{More visualization of data in MVPBench.} Examples of \textbf{Similar Video Pairing} tasks.}
    \label{fig:7}
\end{figure*}

In Figure \ref{fig:5}, Figure \ref{fig:6} and Figure \ref{fig:7}, we present additional visualizations of the data from MVPBench, encompassing the specific prompt designs and the anticipated model responses for each query. Each video in the example is sequentially numbered, and the video token placeholders \textless video\textgreater are aligned with the video numbers specified in the prompt. The numerical response generated by the model corresponds to the video number representing the correct choice.

The complete evaluation code and dataset will be available at \textit{{\textcolor{magenta}{https://github.com/MVPBench/MVPBench}}}, which will also host any subsequent updates to the dataset.

\bibliographystyle{IEEEtran}
\bibliography{main}

\end{document}